\documentclass[10pt,twocolumn,letterpaper]{article}

\usepackage{iccv}
\usepackage{times}
\usepackage{epsfig}
\usepackage{graphicx}
\usepackage{amsmath}
\usepackage{amssymb}
\usepackage{booktabs}
\usepackage{comment}
\usepackage{multirow}
\usepackage{caption}
\usepackage{subcaption}

\usepackage{pifont}% http://ctan.org/pkg/pifont
\newcommand{\xmark}{\ding{55}}%
\newcommand{\cmark}{\ding{51}}%
\usepackage[breaklinks=true,bookmarks=false]{hyperref}
\usepackage{cleveref}
\crefname{section}{§}{§§}

\iccvfinalcopy % *** Uncomment this line for the final submission

\def\iccvPaperID{****} % *** Enter the ICCV Paper ID here

\newcommand{\modelname}{{\usefont{T1}{ptm}{m}{n}GroundNLQ}}
\ificcvfinal\fi

\begin{document}

%%%%%%%%% TITLE
\title{GroundNLQ @ Ego4D
Natural Language Queries Challenge 2023}

\author{Zhijian Hou$^{1}$\thanks{~This work was done during the first author's internship in MSR Asia.}, Lei Ji$^2$, Difei Gao$^3$, Wanjun Zhong$^{4}$, Kun Yan$^5$, Chao Li$^2$,\\
Wing-Kwong Chan$^1$, Chong-Wah Ngo$^6$, Nan Duan$^2$, Mike Zheng Shou$^3$ \\
    $^1$ City University of Hong Kong \quad 
    $^2$ Microsoft \quad
    $^3$ National University of Singapore     \\
    $^4$ Sun Yat-sen University \quad 
    $^5$ Beihang University \quad 
    $^6$ Singapore Management University  \\
    {\tt \small \{zjhou3-c@my., wkchan@\}cityu.edu.hk; 
     \tt \small \{leiji,clao nanduan\}@microsoft.com;}\\
     {\tt \small \{daniel.difei.gao,      mike.zheng.shou\}@gmail.com; 
      \tt \small  zhongwj25@mail2.sysu.edu.cn; }
      \\
     {\tt \small kunyan@buaa.edu.cn;  
      \tt \small cwngo@smu.edu.sg}\\
}

\begin{comment}
\author{First Author\\
Institution1\\
Institution1 address\\
{\tt\small firstauthor@i1.org}
\and
Second Author\\
Institution2\\
First line of institution2 address\\
{\tt\small secondauthor@i2.org}
}
\end{comment}

%\begin{document}

\maketitle
\ificcvfinal\thispagestyle{empty}\fi

%a novel video-language model to ground a natural language query in a video as the 

%%%%%%%%% ABSTRACT
\begin{abstract}
In this report, we present our champion solution for Ego4D Natural Language Queries (NLQ) Challenge in CVPR 2023. Essentially, to accurately ground in a video, an effective egocentric feature extractor and a powerful grounding model are required. Motivated by this, we leverage a two-stage pre-training strategy to train egocentric feature extractors and the grounding model on video narrations, and further fine-tune the model on annotated data. In addition, we introduce a novel grounding model \modelname, which employs a multi-modal multi-scale grounding module for effective video and text fusion and various temporal intervals, especially for long videos. 
On the blind test set, \modelname~achieves 25.67 and 18.18 for R1@IoU=0.3 and R1@IoU=0.5, respectively, and surpasses all other teams by a noticeable margin. Our code will be released at~\url{https://github.com/houzhijian/GroundNLQ}.
\end{abstract}

\section{Introduction}
The Ego4D~\cite{grauman2022ego4d} NLQ challenge aims to localize a temporal window within a long-form first-person video corresponding to a natural language (NL) question. Existing approaches (summarized in Table \ref{table:literature}) primarily explore two research directions: 1) pre-training a representative egocentric feature extractor using the Ego4D video-narration dataset~\cite{lin2022egocentric,chen2022internvideo}, or 2) deploying a powerful grounding model that processes the interaction between video and text features to predict the relevant temporal interval~\cite{liu2022reler,mo2022simple,hou2022efficient}.

To develop a discriminative egocentric video representation, EgoVLP~\cite{lin2022egocentric} and InternVideo~\cite{chen2022internvideo} have been pre-trained on the Ego4D video-narration dataset, thus becoming go-to egocentric feature extractors. Recently, NaQ~\cite{ramakrishnan2023naq} proposes an effective data augmentation strategy to mitigate the data scarcity problem for egocentric video grounding, achieving substantial performance gains. Inspired by this, we employ a two-stage pre-training pipeline that includes both the feature extractor and model pre-training, followed by fine-tuning the grounding model on annotation data.

\begin{table*}[t]
  \centering
   \caption{Comparison among our approach and previous challenge approaches. Results are reported on the blind test split on the Recall@1 metric. }
    \resizebox{0.95\textwidth}{!}{\begin{tabular}{l|c|c|c|c| c|cc}
    \toprule
    Team name & Video Feature & Text Feature & Grounding model & Long Video Processing& Model Pre-train & IoU=0.3 & IoU=0.5 \\
    \midrule
    EgoVLP~\cite{lin2022egocentric} & EgoVLP & EgoVLP & VSLNet & Sparse Sampling  &    & 10.46 & 6.24 \\
    ReLER~\cite{liu2022reler} & SlowFast+Omnivore+CLIP & CLIP  & ReLER & Sparse Sampling  &    & 12.89 & 8.14 \\
    CONE~\cite{hou2022efficient}  & EgoVLP & EgoVLP/CLIP & CONE  &  Window Slicing  &   & 15.26 & 9.24 \\
    Badgers@UW.~\cite{mo2022simple} & SlowFast+Omnivore+EgoVLP & CLIP  & ActionFormer+AdaAttN &  full &     & 15.71 & 9.57 \\
    InternVideo~\cite{chen2022internvideo} & EgoVLP+InternVideo & EgoVLP & VSLNet &  Sparse Sampling   &  & 16.46 & 10.06 \\
    NaQ++~\cite{ramakrishnan2023naq}   & EgoVLP+InternVideo & CLIP  & ReLER & Sparse Sampling &\checkmark & 21.70  & 13.46 \\
    \midrule
    GroundNLQ & EgoVLP+InternVideo & CLIP  & GroundNLQ & full & \checkmark & 25.67 & 18.18 \\
    \bottomrule
    \end{tabular}%
    }
  \label{table:literature}%
\vspace{-0.1in}
\end{table*}%

The long video input presents another challenge for this task. Existing models~\cite{zhang2020span} are primarily designed to handle shorter videos. To address long videos, related works attempted to adapt literature models by sparse sampling~\cite{grauman2022ego4d,liu2022reler} or window-slicing~\cite{hou2022efficient}. However, these approaches lead to information loss or insufficient global contextual encoding. To input the complete video at once, the model of the Badger team~\cite{mo2022simple} re-purposes ActionFormer~\cite{zhang2022actionformer} for video grounding, which integrates local self-attention to extract a feature pyramid from an input video. However, they adopt a late-stage multi-modal fusion network, which results in ineffective multi-modal interaction. Motivated by this, we introduce GroundNLQ, which incorporates well-designed multi-modal multi-scale modules. Our module integrates the textual query and long video deeply in the early stage and constructs the text-aware video feature pyramid to capture temporal intervals of various lengths.

Through the two-stage pre-training pipeline for both the egocentric video feature and the grounding model, our single model \modelname~and final ensemble submission surpasses all other teams by a noticeable margin regarding every evaluation metric (in Table~\ref{tab:leaderboard}). Notably, we also achieve a sizeable performance boost compared with the winner approach of the ECCV22 workshop~(i.e., 10.06\%$\xrightarrow{\text{+81\%}}$18.18\% on R1@0.5 in Table~\ref{table:literature}).)

\begin{figure*}[t]
\begin{center}
\includegraphics[width=0.98\textwidth]{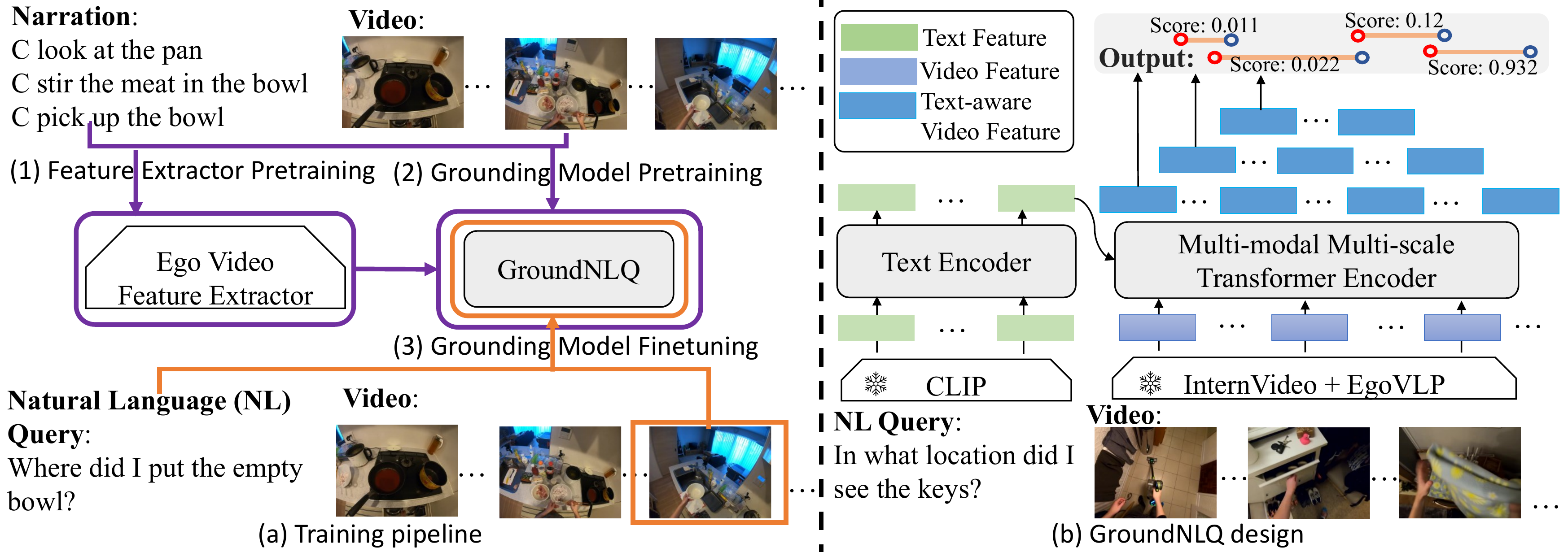}
\end{center}
\caption{Left (a) depicts the overall training pipeline~(\cref{sec:train}). Right (b) shows our grounding model~\modelname~(\cref{sec:model}).}
\label{fig:overview}
\vspace{-0.1in}
\end{figure*}

\section{Methodology}
As depicted in Figure \ref{fig:overview}, we implement a two-stage pre-training process using Ego4D narrations, followed by fine-tuning the model on the annotated dataset. Additionally, we design a novel grounding model, \modelname. This section elaborates on the specifics of both the features (\cref{sec:feature}) and the grounding model (\cref{sec:model}), as well as the implementation details of the training pipeline (\cref{sec:train}).

\subsection{Text and Video Representations}
\label{sec:feature}
For video features, we employ video feature extractors that have been pre-trained on Ego4D narrations, namely, InternVideo~\cite{chen2022internvideo} and EgoVLP~\cite{lin2022egocentric}. We concatenate these features along the channel dimension to formulate the final video feature, mirroring the approach used in~\cite{chen2022internvideo}. Specifically, we extract each video feature for a short snippet (approximately 0.53s) and feed the long video input sequence into our grounding model. In contrast, models such as VSLNet~\cite{grauman2022ego4d,zhang2020span} and ReLER~\cite{liu2022reler} employ sparse sampling for video features to generate a fixed-length frame sequence (e.g., 128 features in VSLNet). For textual features, we use the CLIP~\cite{radford2021learning} text encoder to extract the textual token feature sequence of the natural language (NL) query.

Additionally, we conduct feature projection to map the video and text features into the same embedding space. Specially, we use 2 layers of the 1D Convolution network to project video features, and 2 layers of the linear network to project text features, respectively. Each layer is also stacked with layer normalization and ReLU activation.  After the projection, we add the sin-cos position embedding as in~\cite{devlin2019bert} into the projected video features to embed temporal-aware position information.

% Table generated by Excel2LaTeX from sheet 'Sheet2'
\begin{table*}[t]
  \centering
  \caption { Performance comparison of the top three teams on the public leaderboard. For our submission, we report the performances of both validation and blind test splits.}
    \begin{tabular}{l|cccc|cccc}
    \toprule
    \multicolumn{1}{c|}{\multirow{2}[2]{*}{Team-name}} & \multicolumn{4}{c|}{Validation} & \multicolumn{4}{c}{Test Private} \\
          &  R1 @0.3 & R1 @0.5 & R5 @0.3 & R5 @0.5 &  R1 @0.3 & R1 @0.5 & R5 @0.3 & R5 @0.5 \\
    \midrule
    ego-env &  - & - & - & - & 23.28 & 14.36 & 27.25 & 17.58 \\
    \midrule
    asl\_nlq & - & - & - & - & 24.13 & 15.46 & 34.37 & 23.18  \\
    \midrule
    GroundNLQ* & 26.63 & 18.04 & 52.46 & 40.00    & 24.02 & 16.91 & 40.23 & 28.67 \\
    GroundNLQ & 26.98 & 18.83 & 53.56 & 40.00    & 24.50  & 17.31 & 40.46 & 29.17 \\
    Ensemble & 27.20  & 18.91 & 54.42 & 39.98 & 25.67 & 18.18 & 42.05 & 29.80 \\
    \bottomrule
    \end{tabular}%
  \label{tab:leaderboard}%
\vspace{-0.1in}
\end{table*}%
%for the NLQ Challenge of the Ego4D workshop in CVPR23

\subsection{GroundNLQ Model}
\label{sec:model}
\paragraph{Multi-modal Transformer Encoder.}
For the text encoder, we stack 4 vanilla transformer encoder blocks to learn the contextual token features. Each text transformer encoder block incorporates the multi-head self-attention~(MHA) layer and feed-forward network~(FFN).  

Regarding the video encoder, we use another stack of four transformer encoder blocks to learn text-aware video features. Each video transformer encoder block consists of a local MHA layer, a cross-modal attention layer, and an FFN. 
We leverage window-based local self-attention to decrease the high computational burden of long video modeling, as demonstrated in previous studies~\cite{zhang2022actionformer,beltagy2020longformer}. This approach efficiently handles long sequence inputs while significantly reducing computation costs and maintaining comparable performance.
Additionally, we integrate cross-modal attention to infuse text information into the video features, enabling the early-stage fusion of text and video features. This contrasts with the models by the Badgers team~\cite{mo2022simple}, which employ late-stage fusion following uni-modal feature encoding. This early-stage fusion is critical, given the inherent challenge of aligning video content with textual queries; it ensures comprehensive learning and fusion of multi-modal information.

\paragraph{Multi-scale Transformer Encoder.}
We apply a stack of 6 multi-scale transformer encoder blocks to learn the text-aware video feature pyramid from the text-aware video features. Each block within the multi-scale transformer encoder consists of an MHA layer, a max-pooling layer, and an FFN. We use a max-pooling operator with a stride of 2 to downsample the features, facilitating the capture of longer intervals. The feature pyramid output is a combined result of all 6 layers' outputs with the inputs. This produces 7 text-aware video feature sequence levels of varying lengths for moment prediction. Contrary to the approach in ReLER~\cite{liu2022reler}, which slices the video input sequence into different numbers of clip segments using varying window lengths, our multi-scale mechanism does not employ splitting and is more akin to the feature pyramid network ~\cite{lin2017feature}.

\paragraph{Prediction Head and Loss Function.}
Lastly, we implement two layers of a 1D Convolution network for the classification and regression heads, respectively, following a similar approach as ~\cite{mo2022simple,zhang2022actionformer}. The classification head outputs a probability score for each interval feature of the pyramid, while the regression head outputs the boundary distances from the current interval. The model has dual learning objectives: background/foreground classification and boundary regression. The loss function is a sum of the binary classification loss and IoU regression losses.

\subsection{Implementation Details}
\label{sec:train}

\paragraph{Egocentric Video Feature Extractor Pre-training.}
EgoVLP~\cite{lin2022egocentric} undergoes pre-training with 3.8M paired egocentric video clips and corresponding narrations (i.e., EgoClip~\cite{lin2022egocentric}) via a CLIP-like contrastive loss. Despite a significantly smaller amount of dataset than the 400M image-text pairs in CLIP or 136M video clip-text pairs in Howto100M, EgoVLP demonstrates the necessity of egocentric data pre-training for egocentric video tasks. Additionally, InternVideo~\cite{chen2022internvideo} translates textual narrations into corresponding verb and noun class labels for each EgoClip video clip, utilizing the VideoMAE~\cite{tong2022videomae} backbone to train the 1-of-K classification for separate verb and action labels. The result is three video features pre-trained on Ego4D narrations: EgoVLP, InternVideo-Verb, and InternVideo-Noun. We concatenate these features along the channel dimension for the grounding model's video input.

\paragraph{Grounding Model Pre-training.}
NaQ~\cite{ramakrishnan2023naq} presents a data augmentation strategy that converts standard Ego4D video-text narrations into training data for the grounding model. We adhere to NaQ's data collection steps, assembling all training videos for episode memory benchmarks and collating corresponding $<$video, narration, moment$>$ tuples. Ground-truth temporal boundaries are initialized using the EgoClip boundary and refined via the temporal response jittering strategy~\cite{ramakrishnan2023naq}. We then use this pre-training data to train our model with our training loss function. The training utilizes 4 V100 GPUs with a batch size of 4 per GPU, lasting approximately 4 days. The total and warmup epochs are 10 and 4, respectively, with a maximum learning rate of 2e-4. The best-performing model epoch is chosen based on inference on the validation split in zero-shot mode.

\paragraph{Grounding Model Fine-tuning.}
The fine-tuning stage involves initializing the prediction head from scratch due to moment boundary discrepancies between pre-training and training data. We initially train our model on the train split, determine the optimal epoch number, and report results on the validation split. For leaderboard submission, we train on the combined train+val splits and apply the model with the optimal epoch number to the private test split. This stage uses 2 V100 GPUs with a batch size of 2 per GPU, taking about 8 hours. The total and warmup epochs are 10 and 4, respectively, with a maximum learning rate of 1e-4.

\paragraph{Model Ensemble}
We also explore a model variant called \modelname$\star$, wherein the primary modification is the integration of the cross-modal layer with the multi-scale transformer encoder. This new multi-scale block comprises the MHA layer, the max-pooling layer, the cross-modal layer, and the FFN. For leaderboard submission, we ultimately ensemble the predictions from two models (i.e., \modelname~and \modelname$\star$).

\section{Experiment}

\subsection{Data Analysis}

\begin{table}[htbp]
\footnotesize
\caption{Statistics showing the number of videos and textual narrations/queries in different splits. PT is short for Pre-Train.}
\label{table:dataset_statistics}
\centering
\resizebox{0.48\textwidth}{!}{
\begin{tabular}{c | c  c c c c}
\toprule
Type & Feature PT & Model PT & Train & Validation & Blind Test  \\
\midrule
Video & 3.8M & 5,130 & 1,271 & 415 & 333 \\
\midrule
%\cline{2-6}
Text & 3.8M & 899,219 &  13,849  & 4,552 & 4,005   \\
\bottomrule
\end{tabular}
}
\vspace{-0.1in}
\end{table}

\begin{comment}
\toprule
Dataset & Type & Pre-train & Train & Validation & Test Private  \\
\midrule
\multirow{2}{2em}{Ego4D-NLQ} & video & 5,130 &1,271 & 415 & 333 \\
\cline{2-6}
& query & 899,219 &  13,849  & 4,552 & 4,005   \\
\end{comment}
Table~\ref{table:dataset_statistics} presents the dataset statistics for training the model. For feature pre-training, Ego4D videos are segmented into short clips to learn video representation. For model training, we employ original long videos to train the grounding model.

\subsection{Result Analysis}
Table~\ref{tab:leaderboard} displays our primary leaderboard results. Our best leaderboard submission originates from the ensembled model. Significantly, our results outpace all other teams by a substantial margin across all metrics, particularly the R5 metrics. Furthermore, our single model also exhibits robust performance, thereby demonstrating the superior efficacy of \modelname~and the staged training pipeline.

\subsection{Ablation Analysis}
\begin{table}[htbp]
\caption{Ablations study for InternVideo video feature and model pre-training stage. Results are reported on the validation split. }
\label{tab:ablation}
\begin{subtable}[c]{.9 \linewidth}
\centering
\resizebox{\linewidth}{!}{
    \begin{tabular}{c|cccc}
    \toprule
    InternVideo &  R1@0.3 & R1@0.5 & R5@0.3 & R5@0.5 \\
    \midrule
    \xmark & 21.81 & 14.28 & 45.56 & 32.54 \\
    \cmark & 26.98 & 18.83 & 53.56 & 40.00 \\
    \bottomrule
    \end{tabular}%
}
\caption{Effects of the InternVideo video feature.}
\label{tab:internvideo}
\end{subtable}
\\
\begin{subtable}[c]{.9 \linewidth}
\centering
 \resizebox{\linewidth}{!}{
\begin{tabular}{c|cccc}
    \toprule
    Pre-train &  R1@0.3 & R1@0.5 & R5@0.3 & R5@0.5 \\
    \midrule
    \xmark & 16.74 & 11.47 & 39.02 & 27.39 \\
    \cmark & 26.98 & 18.83 & 53.56 & 40.00  \\
    \bottomrule
    \end{tabular}%
}%
\caption{Effects of the model pre-training stage.}
\label{tab:pre-train}
\end{subtable}
\vspace{-0.1in}
\end{table}

Table~\ref{tab:ablation} provides an ablation study on both features and models. Table~\ref{tab:internvideo} underlines the importance of all features. Using the EgoVLP feature alone, the R1@0.3 performance declines from 26.98\% to 21.81\%. This noticeable gap leads to diminished performance in further ensembles.
Additionally, we explore the fine-grained ranking in CONE~\cite{hou2022efficient}. 
We employ the top5 prediction of the \modelname~model and re-rank these predictions based on the matching score of the EgoVLP model.
This also results in a drop in R1@0.3 performance from 26.98\% to 20.52\%, and the further ensemble does not enhance performance. 
Table\ref{tab:pre-train} emphasizes the importance of the pre-training stage. As found in NaQ~\cite{ramakrishnan2023naq}, pre-training the grounding model substantially enhances performance.

\begin{figure}[t]
% \footnotesize
\begin{center}
\includegraphics[width=0.48\textwidth]{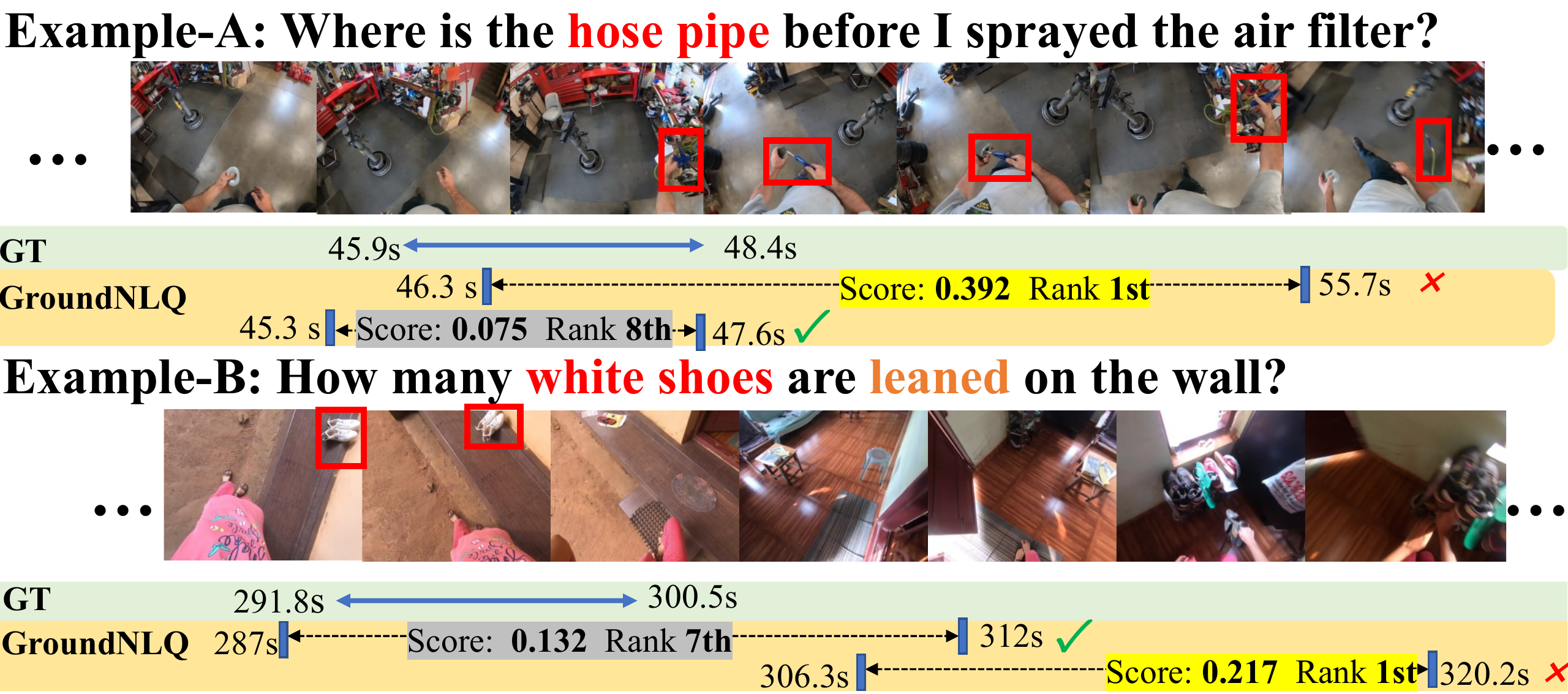}
\end{center}
\caption{Two failure cases. We show the groundtruth segments and our prediction segments.
}
\label{fig:qualitative}
\vspace{-0.1in}
\end{figure}%*

\subsection{Limitation Analysis}
Figure~\ref{fig:qualitative} shows two failure examples contrasting the groundtruth and predicted segments.
In Example-A, the error arises from the imprecise boundary. 
The first prediction of \modelname~captures not only the ground truth event (``I pick up the hose pipe") but also subsequent events (``I spray the air filter with the hose pipe and then put back the hose pipe"). Consequently, the matched IoU with the ground truth is below the 0.3 threshold, classifying the prediction as false.
In Example-B, the error results from an inadequate understanding of the video content and textual query's nuances. The first prediction of \modelname~identifies the event where "I enter the room, take off the shoes and put the shoes in the shoebox near the wall". While the key objects (shoes and wall) are matched, the alignment neglects detailed attributes of the shoes (their color is white, and they lean against the wall).

\section{Conclusion}
We present our solution to the Ego4D natural language queries challenge in CVPR 2023. Through our experiments, we highlight the importance of the two-stage pre-training of both the video feature extractor and the grounding model. Regarding the model, we find the importance of early-stage fusion between long video and textual query in the multi-modal multi-scale module. 
Moreover, we also identify further challenges of the imprecise boundary and fine-grained attribute understanding issues for future work.

{\small
\bibliographystyle{unsrt}
\bibliography{egbib}
}

\appendix

\end{document}